\documentclass[10pt,twocolumn,letterpaper]{article}

\usepackage{iccv}
\usepackage{times}
\usepackage{epsfig}
\usepackage{graphicx}
\usepackage{amsmath}
\usepackage{amssymb}
\usepackage{caption}

\def\iccvPaperID{4011} % *** Enter the ICCV Paper ID here

\ificcvfinal\fi

\usepackage{pifont}
\newcommand{\cmark}{\ding{51}}%
\newcommand{\xmark}{\ding{55}}%

\usepackage[pagebackref=true,breaklinks=true,letterpaper=true,colorlinks,bookmarks=false]{hyperref}

\newcommand{\figureref}[1]{Fig.~\ref{#1}}

\renewcommand{\paragraph}[1]{\vspace{1mm}\noindent\textbf{#1}}

\iccvfinalcopy % *** Uncomment this line for the final submission

\begin{document}

%%%%%%%%% TITLE
% \title{Towards Telepresence on Every Desk}
\title{Bringing Telepresence to Every Desk}

\author{Shengze Wang \quad Ziheng Wang \quad Ryan Schmelzle \quad Liujie Zheng \\ YoungJoong Kwon \quad Soumyadip Sengupta \quad Henry Fuchs\\\vspace{-1em}
\\\vspace{-1.5em}
UNC Chapel Hill\\
\\\vspace{-0.5em}
{\tt\small shengzew@cs.unc.edu} \quad 
{\tt\small \{zihengcwang,ryancschmelzle@gmail.com,edwardqzheng\}@gmail.com} \\
{\tt\small \{youngjoong,ronisen,fuchs\}@cs.unc.edu}
}

% \author{Shengze Wang\\
% UNC Chapel Hill\\
% {\tt\small shengzew@cs.unc.edu}
% \and

% \maketitle

%%%%%%%%%%% TEASER
%\twocolumn[\maketitle\vspace{-3em}\input{fig/teaser}\label{fig:teaser}\bigbreak]

% \twocolumn[\maketitle\vspace{-3em}\includegraphics[\textwidth]{images/teaser.pdf}\bigbreak]

% \twocolumn[\vspace{-1cm}\maketitle\vspace{-3em}\input{images/teaser.pdf}\bigbreak]

\twocolumn[{%
\renewcommand\twocolumn[1][]{#1}%
\vspace{-10mm}
\maketitle
\vspace{-5mm}
\begin{center}
\vspace{-5mm}
    \centering
    \captionsetup{type=figure}
    \includegraphics[width=.97\textwidth]{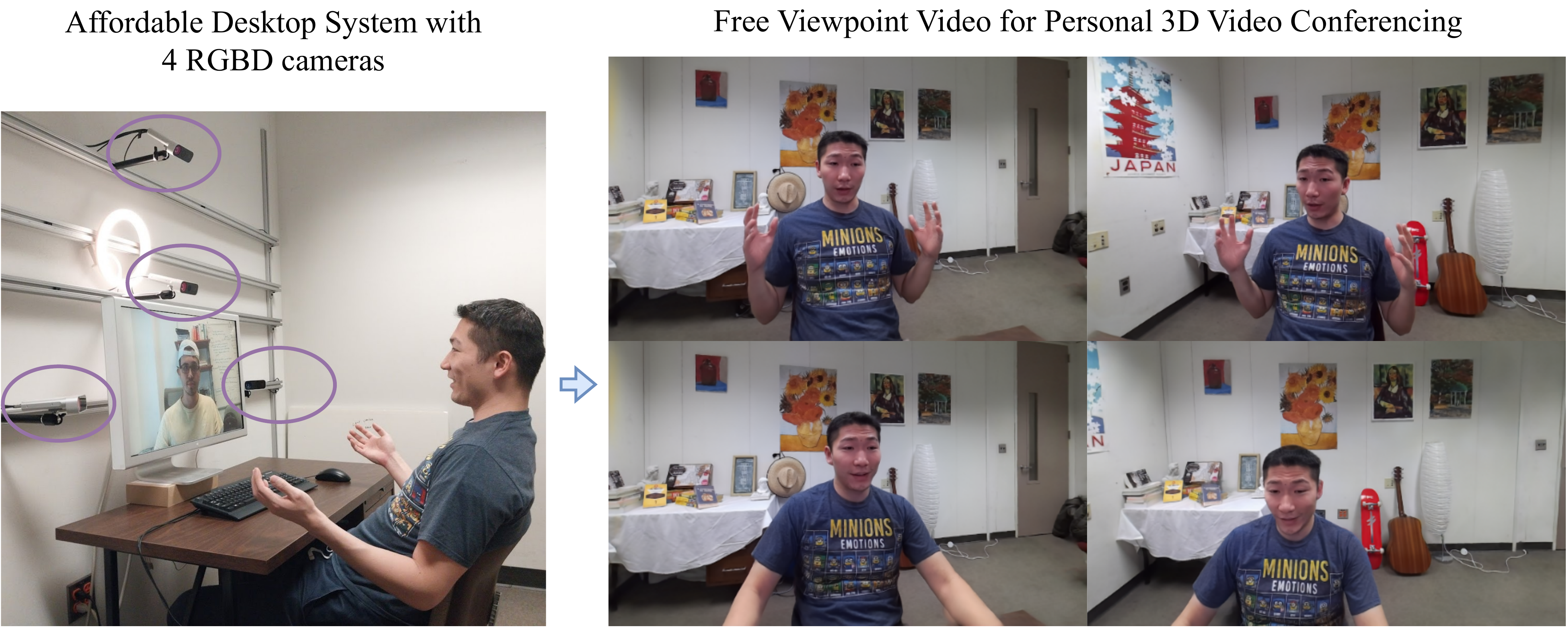}
    \captionof{figure}{\small{We present a capturing and rendering system designed for personal telepresence. (left) Our system utilizes four RGBD cameras to render high-resolution free-
    viewpoint videos, which are crucial to immersive 3D video conferencing. (right) We show synthesized images from different novel viewpoints across different frames.}}
    \label{fig:teaser}
\end{center}%
\vspace{0em}
}]

%%%%%%%%% ABSTRACT

\begin{abstract}
\vspace{-3mm}
In this paper, we work to bring telepresence to every desktop. Unlike commercial systems, personal 3D video conferencing systems must render high-quality videos while remaining financially and computationally viable for the average consumer. To this end, we introduce a capturing and rendering system that only requires 4 consumer-grade RGBD cameras and synthesizes high-quality free-viewpoint videos of users as well as their environments.

Experimental results show that our system renders high-quality free-viewpoint videos without using object templates or heavy pre-processing. While not real-time, our system is fast and does not require per-video optimizations. 
Moreover, our system is robust to complex hand gestures and clothing, and it can generalize to new users.
This work provides a strong basis for further optimization, and it will help bring telepresence to every desk in the near future.
The code and dataset will be made available on our website \url{https://mcmvmc.github.io/PersonalTelepresence/}.

\end{abstract}

\vspace{-0.5em}
\section{Introduction} \label{sec:intro}

\vspace{-0.5em}

In recent years, video conferencing has become ubiquitous in our daily lives, facilitated by software such as Zoom, Gather, and Meet. However, 2D video conferencing fails to provide immersive experiences and realistic conversations. Thus, researchers have worked for decades to develop 3D telepresence systems that enable remote users to virtually share the same space~\cite{officeoffuture,draper1998telepresence,gibbs1999teleport,kuster2012towards,maimone2012enhanced}.

Unlike systems using cumbersome headsets~\cite{holoportation}, encumbrance-free systems~\cite{starline,virtualcube,maimone2012enhanced,officeoffuture} leverage regular/autostereo displays to provide immersive experiences without requiring the users to wear headsets. Recent examples such as Project Starline~\cite{starline} and VirtualCube~\cite{virtualcube} have achieved unprecedented levels of realism. However, these commercial systems require intricate hardware setups and dedicated physical spaces (\ie booths or rooms) making them inaccessible to the general public.

Different from commercial systems, personal telepresence must be affordable and accessible to a wider range of users. Our system achieves this through the use of sparse and consumer-quality sensors. Specifically, our system employs only 4 Microsoft Azure Kinect RGBD cameras, which cost approximately \$1600 in total - less than half the price of AR glasses such as Magic Leap and Hololens, which can cost over \$3500. To use our system, a user only needs to install the four cameras around their monitor, without requiring any additional hardware or dedicated capture spaces/booths. This setup provides two benefits: first, it enhances the sense of realism by placing the users in the context of their actual surroundings, creating a more immersive telepresence experience and contributing to a sense of personal connections between users. Second, this lightweight system reduces the financial and physical burden on consumers.  However, this lightweight setup also presents several challenges:

\begin{enumerate}
\vspace{-2mm}

\item \textit{Sparse viewpoints}: 
The sparsity of viewpoints results in wider baselines and larger perspective changes, making accurate reconstructions more challenging. 
\vspace{-2mm}

\item \textit{Inaccurate and noisy depth sensors}: Depth measurements from RGBD cameras are biased, inducing alignment errors (Fig.\ref{fig:sweep_compare}). Moreover, depth values vary across frames, resulting in flickering.

\vspace{-2mm}
\item \textit{Background}: Without dedicated booths, the system needs to synthesize high-quality renderings of the background in addition to the foreground.

\end{enumerate}

\vspace{-2mm}
To this end, we propose a relatively capturing and rendering system. Our system synthesizes high-quality novel view images of human subjects and backgrounds given 4 input RGBD streams. To improve the reconstruction quality, we introduce the Multi-layer Point Cloud (MPC), a novel volumetric representation designed for biased depth inputs. MPC is constructed by sweeping point clouds from the input viewpoints. Compared to the conventional novel-view-depth-sweeping, MPC enables more accurate reconstruction of slanted surfaces and contour regions, thus reducing flickering. To further improve temporal smoothness, our temporal neural renderer aggregates information across frames. To achieve high-resolution video synthesis under limited GPU memory, we introduce the Spatial Skip Connection inspired by UNet\cite{unet} skip connections. To study the efficacy of our system, we created a dataset tailored for personal 3D video conferencing, \ie the Personal Telepresence Dataset. Experiments show that our system outperforms baseline methods.  Ablation studies further show that each proposed module improves the stability and accuracy of the results. 

Our work can be summarized as follows:
\begin{itemize}
\vspace{-2mm}

\item We present a relatively affordable capturing and rendering system for desktop telepresence using only 4 RGBD cameras. Our setup can be easily replicated on any desk without a dedicated space or booth.
\vspace{-2mm}
\item We designed the Multi-Layer Point Cloud (MPC), a new volumetric representation that improves reconstruction from RGBD inputs. We further improve the rendering stability and memory efficiency via our temporal renderer and Spatial Skip Connections.

\vspace{-2mm}
\item Experiments show that our system outperforms recent competitive methods in the synthesis of free-viewpoint videos. Our method is fast and does not require object templates or heavy pre-processing. Moreover, it is robust to complex hand gestures and clothing, and it can generalize to new users.
\end{itemize}

%
%%%%%%%%%%%%%%%%%%%%%%%%%%%%
%
\begin{table*}[t]
\centering
\caption{\textbf{Prior Telepresence Systems}:  A high-level comparison between our system and prior approaches. 
Features can theoretically be implemented but have not yet been shown in prior works; these features are indicated with \emph{*}
}
 \resizebox{0.8\textwidth}{!}{
\begin{tabular}{ l c c c c c c c}
    \hline
    Systems & \multicolumn{1}{l}{Desktop} & \multicolumn{1}{l}{Affordable} & \multicolumn{1}{l}{Generalizable} & \multicolumn{1}{l}{Background} & \multicolumn{1}{l}{Quality} & \multicolumn{1}{l}{Real-Time} & \multicolumn{1}{l}{Complexity} \\
    \hline
    Maimone \etal \cite{maimone2013} & \cmark & \cmark & Yes & Yes & Low & \cmark & Low\\
    \hline
    Holoportation \cite{holoportation} & \xmark & \xmark & Yes & \emph{*} & Medium & \cmark & High\\
    \hline
    LookinGood\cite{lookingood} & \xmark & \xmark & \emph{*} & \emph{*} & High & \cmark & High\\
    \hline
    Project Starline\cite{starline} & \xmark & \xmark &Yes & No & High & \cmark & High \\
    \hline
    VirtualCube\cite{virtualcube} & \cmark & \cmark & Yes & \emph{*} & Medium & \cmark & Low \\
    \hline
    Ours & \cmark & \cmark & Yes & Yes & High & 0.19-1.28fps & Low \\
    \hline
  \end{tabular}}
  \vspace{-5mm}
\end{table*}
\vspace{-0.5em}
\section{Related Work} \label{sec:related}

\subsection{3D Video Conferencing Systems}
\vspace{-0.5em}

Since the pioneering works of Cruz-Neira \etal~\cite{thecave} and Raskar\etal~\cite{officeoffuture}), there has been a plethora of work~\cite{animmersive3dconf,maimone2011,maimone2013,towards3dteleconf,viewport,starline,virtualcube} on headset-free 3D video conferencing.

\textbf{Personal Vs. Commercial Systems} The key difference between commercial and personal systems is that personal systems must achieve high-quality results while remaining financially and computationally viable for consumers. As a result, the quality of sensors and complexity of the systems tend to be lower than commercial alternatives. Moreover, recent commercial systems such as Starline~\cite{starline} and VirtualCube~\cite{virtualcube} resemble photo-booths blocking the background. However, in personal usage, it is impractical to dedicate a space specifically for video conferencing. Moreover, the background is often crucial to creating realistic, informal, and intimate personal interactions. Therefore, we envision personal systems that can be installed in a typical room setting and capture background environments.

\subsection{Novel View Synthesis}
\vspace{-0.5em}
\textbf{Dynamic View synthesis} There are many different approaches to synthesizing free-viewpoint videos; Yoon et al.~\cite{nvidiadynamicnvs} utilizes multi-view stereo and monocular depth estimators to generate 3D videos without per-video optimization or prior knowledge of the scene. Additionally, many NeRF-based approaches~\cite{videonerf,nsff,dynamicnerf,nonrigidnerf} encode dynamic scenes as spatiotemporal radiance fields. \cite{videonerf,nsff,dynamicnerf,nonrigidnerf} learn a radiance field and a motion field for each frame. T{\"o}RF~\cite{torf} uses Time-of-Flight sensors to achieve better modeling of both static and dynamic scenes. Approaches like Nerfies~\cite{nerfies}, HyperNeRF~\cite{hypernerf}, and Neural 3D Videos (N3V)~\cite{neural3dvideos} use latent codes to help model dynamic contents. 

However, each of these approaches requires extensive training time. Moreover, all these approaches require buffering the video in advance, making them unsuitable for instant live-streaming applications. 
ENeRF~\cite{enerf} demonstrated notable improvements in generalization over prior works \cite{pixelnerf,mvsnerf,ibrnet,SRF} while achieving real-time rendering without optimization. 
LookinGood~\cite{lookingood} uses multiple RGBD cameras to reconstruct a human mesh in real time and uses a neural network to render the colored mesh from high-quality novel views. While this approach heavily depends on the quality of the preprocessed reconstruction, our approach uses raw RGBD frames to eliminate the need for preprocessing and naturally renders the background and foreground at the same time.

\textbf{Human-Specific Approaches.} Some recent works \cite{neuralbody,neuralactor,anerf,animatablenerf,neuralhumanperformer} exclusively focus on animating clothed humans. They often use colored videos as inputs and leverage human body templates (\eg SMPL\cite{SMPL} and STAR\cite{STAR}) and deep textures. On the other hand, HVSNet \cite{hvsnet} uses monocular RGBD videos to render human subjects from feature point clouds. Our work requires the reconstruction of general scenes including both humans and objects, making these human-only approaches inapplicable. Moreover, our application requires a detailed depiction of the upper body and hands, presenting a challenge for human models designed for portrait or full-body views. Therefore, these approaches would be ill-suited for our application.

\textbf{Preprocessed 3D Representations.} Some works \cite{FVS,SVS,PBNRPO,ADOP,npbg} rely on raw 3D geometry (\eg point clouds and meshes) generated from multi-view stereo software such as COLMAP\cite{colmap}. Such approaches are less prone to generating fog-like artifacts common in NeRF-based methods. Methods like \cite{FVS,SVS} also demonstrate good generalization to new scenes. Point cloud-based neural rendering approaches \cite{PBNRPO,ADOP,wiles2020synsin} show impressive sharpness and details in large scenes with thin structures. However, these approaches either require lengthy training for each frame or do not show competitive results. 

\begin{figure*}
\vspace{-3mm}
  \centering
  \includegraphics[width=0.95\textwidth]{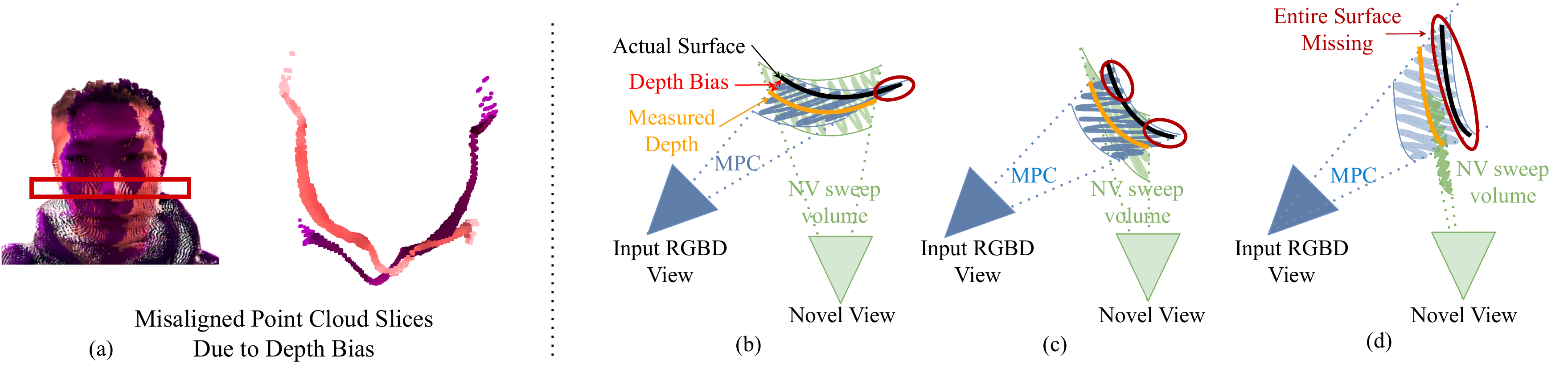}
   \caption{\textbf{MPC Volumes Vs. Conventional Novel View Sweep Volumes.} (a) Depth measurements from RGBD cameras are biased (\ie offset from the true geometry), leading to misalignment between point clouds from different views. (b) Sweeping from the novel view (green volumes) may miss the true surfaces (identified by red circles). The error worsens as the surface gets more slanted in the novel view as shown in (c) and (d). In contrast, our Multi-layer Point Cloud (blue volume) better covers the true geometry, leading to better reconstruction.}
  \label{fig:sweep_compare}
  \vspace{-0.5cm}
\end{figure*}

\begin{figure}
  \centering
  \includegraphics[width=0.48\textwidth]{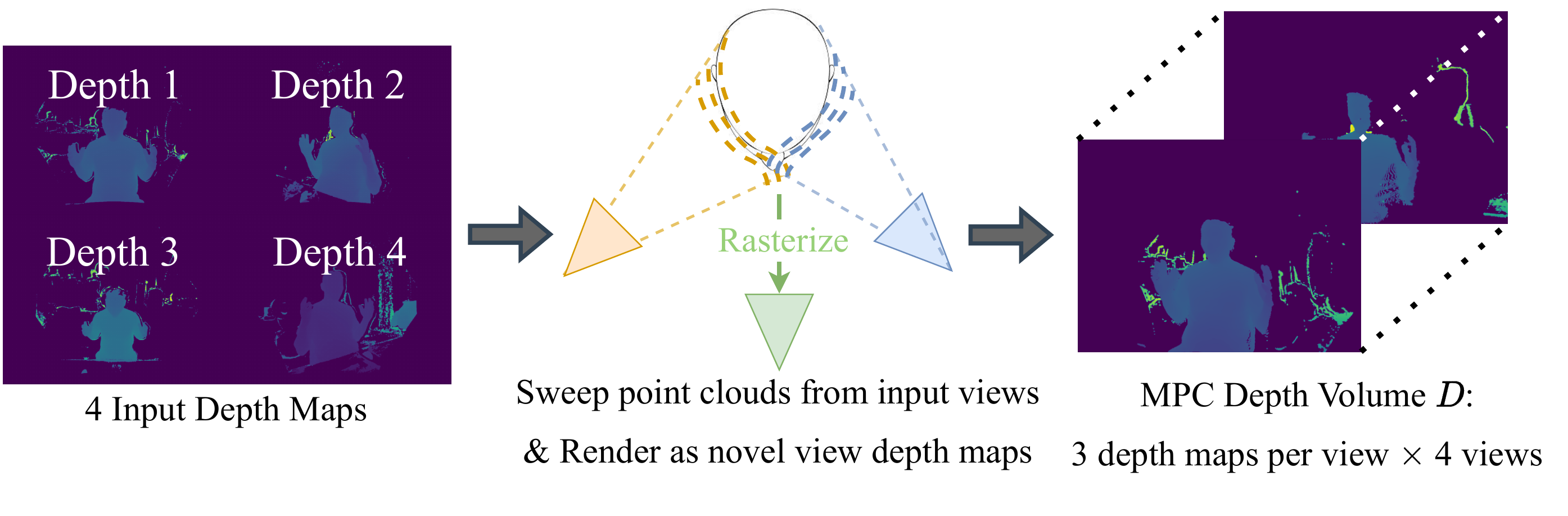}
  \vspace{-8mm}
  \caption{\textbf{Construction of Multi-layer Point Cloud (MPC) Volume}. Each of the $K=4$ input depth maps is perturbed/swept with $N=3$ offset values. The $N$ perturbed depth maps are then lifted into point clouds and rasterized into depth maps in the novel view, producing the MPC depth volume $D \in \mathbb{R}^{K\times N \times H \times W}$ that stores $K\times N$ depth candidates for each pixel in the novel view.}
  \label{fig:MPC_construction}
  \vspace{-5mm}
\end{figure}

\vspace{-0.5em}
\section{Motivation for Multi-Layer Point Cloud}\label{sec:volumes}
In this section, we discuss the motivation for our novel volumetric representation, \ie, Multi-layer Point Cloud.

\textbf{Conventional Novel View Sweep Volume.} 
Many RGB-only approaches~\cite{mvsnerf,mvsnet,casmvsnet,ucsnet,pixelnerf,enerf,deepview,stereomag,DeepMVS,planesweepcollins,planesweepmarc} estimate scene geometry by depth-sweeping from the novel view camera. This approach is proven effective for RGB-only scenarios and is thus naturally extensible to RGBD-based novel view synthesis. For example, VirtualCube~\cite{virtualcube} first generates an averaged novel view depth map from input depth maps. It then generates depth candidates via a novel view depth-sweep around the averaged depth map. However, we show that this simple extension can cause inaccurate reconstructions due to biases in depth sensors. 

\textbf{Challenges Induced by Depth Bias.} Modern commodity depth sensors (\eg Microsoft Azure Kinect) can now achieve good quality but still suffer from noise and bias. Depth bias is especially important because it can cause notable misalignment between views. Fig. \ref{fig:sweep_compare}(a) shows an example of such misalignment between point clouds from two cameras, implying that the depth values are offset from the true geometry.
While conventional volumes can locally search for the true surfaces, their sweep/search volumes might not cover the true surface. As illustrated by Fig. \ref{fig:sweep_compare}(b)-(d), the direction of depth bias lies along the viewing direction of the input camera, but the depth sweep is in the novel view direction. As a result, the sweep volume (green) inevitably misses the true surface, leading to incomplete reconstructions. This effect worsens as the surface becomes more slanted in the novel view (Fig.\ref{fig:sweep_compare}(c)-(d)). Moreover, the sweep volume (green) might completely miss the true surface when the surface is too steep in the novel view (Fig.\ref{fig:sweep_compare}(d)). This problem is amplified by the unstable depth values that vary between frames, inducing flickering around object contours and steep surfaces. 
As a result, previous methods require careful post-processing to stabilize generated videos (\eg averaging adjacent frames~\cite{virtualcube,lookingood}). 

\textbf{Multi-layer Point Cloud (MPC).} 
Depth biases are difficult to remove because they depend on a wide range of factors~\cite{kinectbias1,kinectbias2,kinectbias3,toflightingbias,tofmaterialdistortion}: the viewing angles, materials, and distances to the captured surfaces, the strength and color of lighting, etc. Therefore, to address the aforementioned issues, we designed the Multi-layer Point Cloud, a new volumetric representation more suitable for RGBD inputs. Contrary to conventional novel view sweep volumes, MPC volumes are generated from the input views instead of the novel view. Therefore, MPC volumes are always aligned with the direction of depth biases. As a result, MPC volumes (blue, Fig. \ref{fig:sweep_compare}(b)-(d)) ensure better coverage of true surfaces despite the steep viewing angles and can improve the reconstruction of slanted surfaces (\eg object contours). More algorithmic details are in Sec.~\ref{sec:MPC}.
\vspace{-2mm}
\section{Method}\label{sec:method}
Our goal is to render stable and photorealistic free-viewpoint videos for personal telepresence using a few (\eg 4) RGBD streams. In Sec. \ref{sec:hardware}, we describe our capture system. In Sec. \ref{sec:rendering_sys}, we describe our rendering system.

\subsection{
Personal Desktop Capture System.}\label{sec:hardware}
\vspace{-0.5em}
Our capture system is shown in Figure. \ref{fig:teaser}. We enhance the typical video conferencing setup (\ie one camera mounted near the screen) by placing 4 Microsoft Azure Kinect RGBD cameras near a 23-inch display. We place 3 of the cameras to the left, right, and above the display to provide complete coverage of the user, and the 4th camera right above the monitor to provide details. The cameras are hardware-synchronized via audio cables. Each camera captures color images at 2048$\times$1536 and depth images, which are later merged into a single RGBD image via reprojection and rasterization. Additionally, a ring light is placed behind the middle camera in order to reduce shadows and improve the visibility of facial details. The Azure Kinect cameras are priced at \$399 each (totaling \$1596 for 4 cameras), similar to high-end VR headsets like Meta Quest Pro (\$1500) and cheaper than typical AR glasses (\eg Magic Leap and HoloLens at over \$3500).

\begin{figure*}
  \centering
  \vspace{-2mm}
  \includegraphics[width=0.95\textwidth]{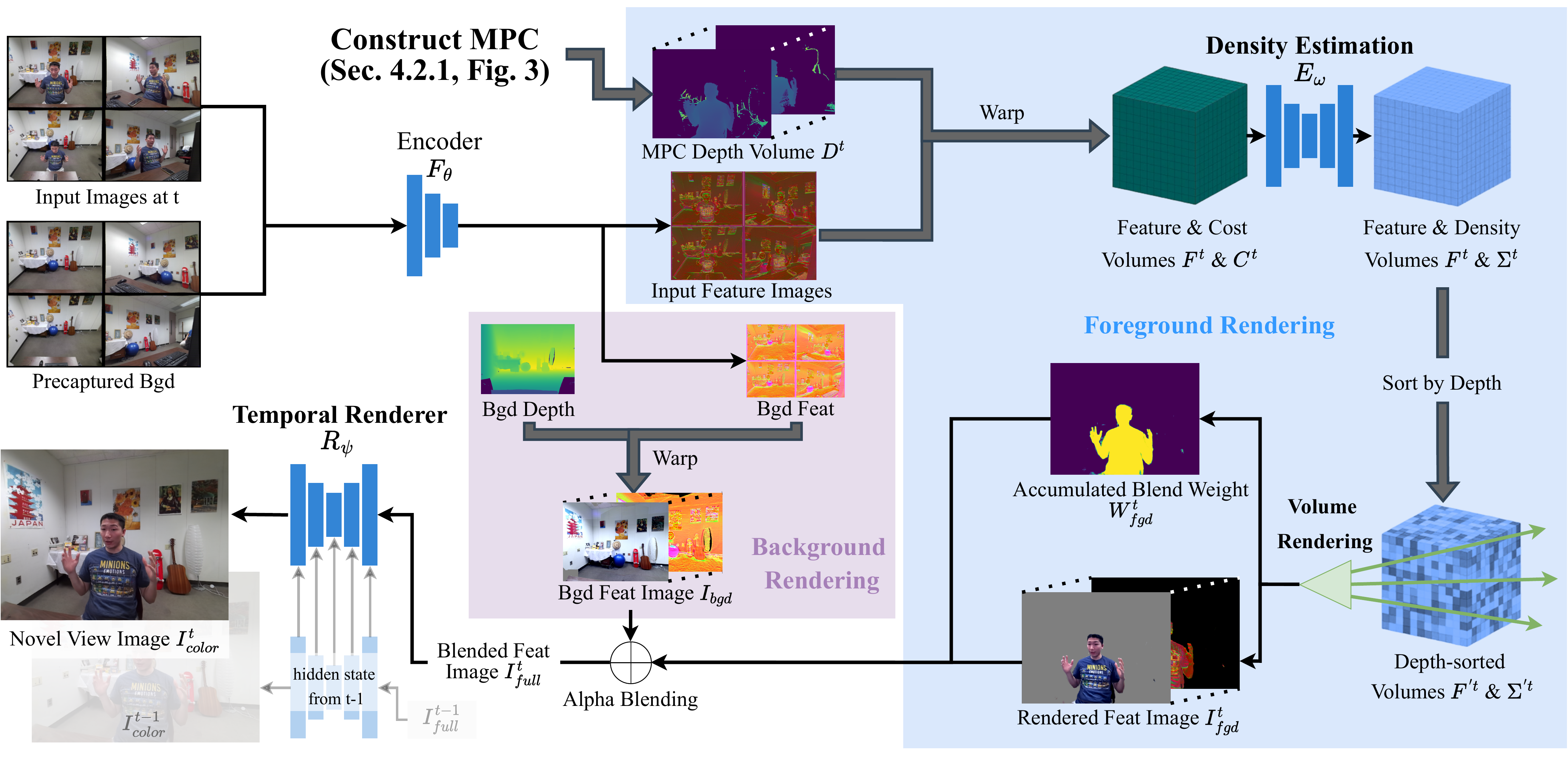}
  \vspace{-4mm}
   \caption{\textbf{System Overview:} (1) Foreground (Blue): Given the 4 input RGBD images at frame $t$, we first construct an MPC depth volume $D^t$. $D^t$ is used to warp the input feature images to the novel view and construct the feature volume $F^t$ and the cost volume $C^t$. Given $F^t$ and $C^t$, A 3D ConvNet $E_\psi$ then estimates a density volume $\Sigma^t$. To enable volume rendering, volumes $\Sigma^t$ and $F^t$ are then sorted by depth based on $D^t$, producing $F'^t$ and $\Sigma'^t$. Volume rendering on $F'^t$ thus produces the feature image $I_{fgd}^t$ and the accumulated blend weight map $W^t_{fgd}$. (2) Foreground+Background(Purple): using the novel view background depth, we warp and average input feature images of the background into a background feature image $I_{bgd}$. We then alpha blend $I_{bgd}$ and $I_{fgd}^t$ into a full feature image $I_{full}^t$ using $W^t_{fgd}$. Our temporal renderer then leverages the hidden state from $t-1$ to render the novel view image $I_{colot}^t$. 
   }
  \label{fig:overview}
  \vspace{-4mm}
\end{figure*}

\subsection{Rendering System.}\label{sec:rendering_sys}
\vspace{-0.5em}
\figureref{fig:overview} shows the overview of our rendering system. We first describe the general rendering pipeline that utilizes Multi-Layer Point Cloud (Sec. \ref{sec:MPC}). Then, we describe our Temporal Renderer (Sec. \ref{sec:temp_agg}) and Spatial Skip Connection (Sec. \ref{sec:spatial_skip}) that further improve the video quality.

\subsubsection{Constructing the MPC Volume} \label{sec:MPC}
\vspace{-2mm}
As discussed in Sec. \ref{sec:volumes}, MPC improves reconstruction because it is robust to depth biases in RGBD cameras. In Fig. \ref{fig:MPC_construction}, we illustrate the construction of MPC volumes. For each of the $K=4$ input views, we first generate $N=3$ copies of the input depth maps, each perturbed by a value of $\triangle d$; we use $\triangle d\in [-1cm,0cm,+1cm]$ in our experiments. These perturbed input depth maps are then lifted into 3D space, resulting in $N$ layers of point clouds for each of the $K$ views. These point clouds are then rasterized into the novel view as depth maps, \ie the MPC depth volume. Thus the MPC depth volume $D \in \mathbb{R}^{K\times N \times H \times W}$ stores $K\times N$ depth candidates for each pixel in the novel view. Notice that the MPC depth volume is ordered by cameras (\ie the first $N$ depth maps are from view 1, and the next $N$ from view 2, etc.), and thus the $K\times N$ depth candidates for a pixel are not sorted by depth. During our experiments, these depth volumes are preprocessed to avoid redundant computation and accelerate training and testing.

\subsubsection{Density Estimation} \label{sec:density}
\vspace{-0.5em}
The MPC depth volume $D$ provides a set of depth candidates for each pixel in the novel view image. We then estimate density values for each candidate similar to prior works~\cite{mvsnerf,mvsnet,enerf,casmvsnet,ucsnet}. 
More specifically, we lift and project each candidate into all input views. The averaged feature of a candidate across views is stored in the feature volume $F$, and its feature variance across views is stored in the cost volume $C$. Good candidates (\ie close to the true surface) are more likely to have low variances (\ie high multi-view consistency) than bad candidates. 
To account for occlusions, only the unoccluded views (\ie the candidate's projected depth is consistent with the input view's depth map) are used during averaging and variance calculation. Finally, a 3D ConvNet $E_\psi$ estimates density values for each candidate based on the feature volume $F$ and cost volume $C$, producing the density volume $\Sigma$. 
Notice that $F$, $C$ and $\Sigma$ are not sorted by depth because the MPC depth volume is ordered by cameras (Sec. \ref{sec:MPC}).

\subsubsection{Rendering} \label{sec:rendering}
\vspace{-0.5em}
To achieve high-quality results, we leverage volumetric rendering, a key component to recent advances in novel view synthesis~\cite{plenoxels}. Given the depth-sorted density volume $\Sigma'^t$, volume rendering blends all candidates along the ray $r$ of a pixel $p$ into one feature pixel in the image $I_{fgd}^t$:
\vspace{-2mm}
\begin{equation}
\label{eqn:transmittance}
T_i = exp(-\sum_{j=1}^{K\times N -1}\sigma_j)
\end{equation}
\vspace{-3mm}
\begin{equation}
\label{eqn:vol_render}
I^t_{fgd}(r) = \sum_{i=1}^{K\times N}T_i(1-exp(-\sigma_i))f_i)
\end{equation}
$f_i$ is the feature stored in $F'^t$ for the $i$th candidate of pixel $p$. The result is a blended feature image $I^t_{feat}\in \mathbb{R}^{H \times W \times F}$ and an accumulated blend weight map $W_{feat}^t\in \mathbb{R}^{H \times W }$ (please refer to \cite{nerf} for more details). Given the warped and averaged background feature image (Sec. \ref{sec:bgd}), we alpha blend the background and foreground feature images into a full feature image $I_{full}^t$ using $W_{fgd}^t$. $I_{full}^t$ and $W_{fgd}^t$ are then fed to a UNet-based renderer to generate a color image $I_{color}^t$ and a feature image $I_{out}^t$ at time t. 

\subsubsection{Stabilization via Temporal Renderer}
\label{sec:temp_agg}
\vspace{-0.5em}
Our rendering system is capable of generating high-quality images without knowledge of prior frames. However, flickering artifacts between adjacent frames may occur due to the noise in depth measurements from RGBD cameras. Some prior approaches perform stabilization by averaging adjacent frames, e.g., LookinGood~\cite{lookingood} and VirtualCube~\cite{virtualcube}. However, this approach produces ghosting artifacts under fast motions, and flickering persists on steep surfaces. To mitigate these artifacts, we use a recurrent renderer $R_\psi$ that exploits the temporal information between frames to suppress temporal inconsistencies. 

First, our renderer $R_\psi$ (a GRU-UNet) synthesizes the previous frame $I_{color}^{t-1}$ at the viewpoint of interest. During the synthesis, the GRU captures useful temporal information and stores it as its hidden state. Then, the renderer uses the hidden state to condition the synthesis of $I_{color}^{t}$ at the same viewpoint. This formulation enables our renderer to selectively utilize or discard information from the prior frame and synthesize more stable and accurate videos, even for fast motions.

\begin{figure*}
  \centering
  \vspace{-2mm}
  \includegraphics[width=0.99\textwidth]{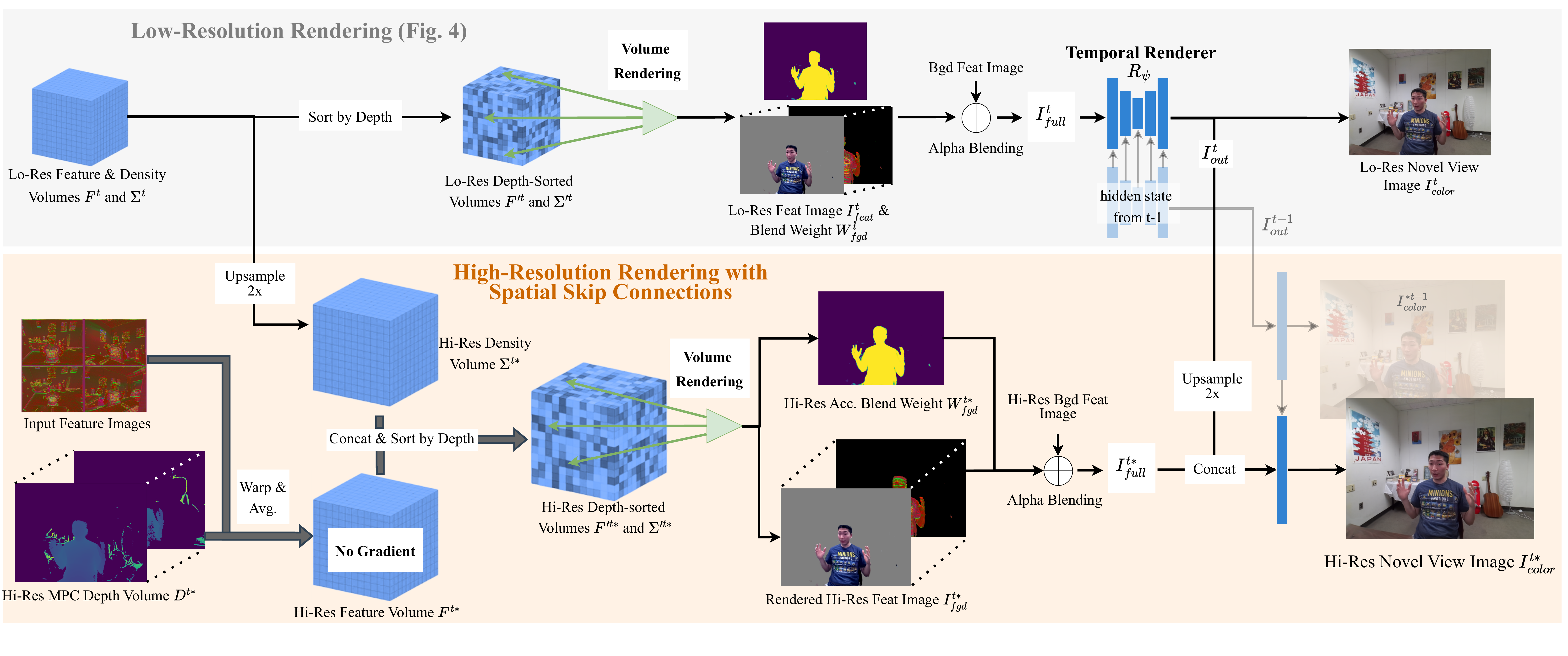}
  \vspace{-4mm}
   \caption{\textbf{Spatial Skip Connections (SSC):} Similar to skip connections in UNet\cite{unet}, SSC directly propagates low-level details to a convolution layer by concatenating a high-resolution feature map $I_{full}^{t*}$ with an upsampled feature map $I^t_{out}$. To calculate $I_{full}^{t*}$, we perform volume rendering using high-resolution feature and density volumes $F^{t*}$ and $\Sigma^{t*}$. To avoid the memory-heavy 3D convolution, $\Sigma^{t*}$ is upsampled from $\Sigma^t$ in order to approximate the actual density volume. Additionally, $F^{t*}$ is constructed without gradient calculation.}
  \label{fig:ssc}
  \vspace{-4mm}
\end{figure*}

\vspace{-0.5em}
\subsubsection{High-Resolution via Spatial Skip Connections}
\label{sec:spatial_skip}
\vspace{-0.5em}
Our rendering system requires 3D convolution for volume rendering, and high-resolution rendering can easily exceed the GPU memory capacity. A straightforward solution to achieve high resolution is to perform convolution on an upsampled low-resolution input. However, this technique relies on convolution layers to hallucinate high-resolution details, which can be unreliable. To overcome this challenge, we propose the Spatial Skip Connection (SSC) to enhance the details. Similar to skip connections in UNet\cite{unet}, SSC directly propagates low-level details to a convolution layer by concatenating a high-resolution feature map $I_{full}^{t*}$ with an upsampled feature map $I^t_{out}$. 

To calculate the skip connection $I_{full}^{t*}$, we perform high-resolution rendering using the method in previous sections, but with two changes that save the memory: (1) Instead of performing costly 3D convolution to estimate a high-resolution density volume, 
we approximate it by upsampling $\Sigma^t$ to $\Sigma^{t*}$. (2) we do not calculate the gradient for the high-resolution feature volume $F^{t*}$. Volume rendering then generates the high-resolution blend weight map $W^{t*}_{fgd}$ and foreground feature image $I^{t*}_{fgd}$. $I^{t*}_{fgd}$ could thus be calculated by alpha blending $I^{t*}_{fgd}$ and the high-resolution background feature image (Sec.~\ref{sec:bgd}). $I^{t*}_{fgd}$ is then concatenated with $I^t_{out}$ (Sec.~\ref{sec:rendering}) before being decoded into the final high-resolution novel view image $I_{color}^{t*}$

\vspace{-0.5em}
\subsubsection{Background Pre-Capture}\label{sec:bgd}
\vspace{-0.5em}
To reduce occlusion and improve stability, we pre-capture the background at the start of each capturing session without the users. The background depth maps are lifted into point clouds and rasterized in the novel view as depth maps. These rasterized background depth maps are then used to warp features from the input views into the novel view in order to generate a background feature image (Sec.~\ref{sec:rendering} and Sec.~\ref{sec:spatial_skip}).

\begin{figure*}
  \centering
  \includegraphics[width=\textwidth]{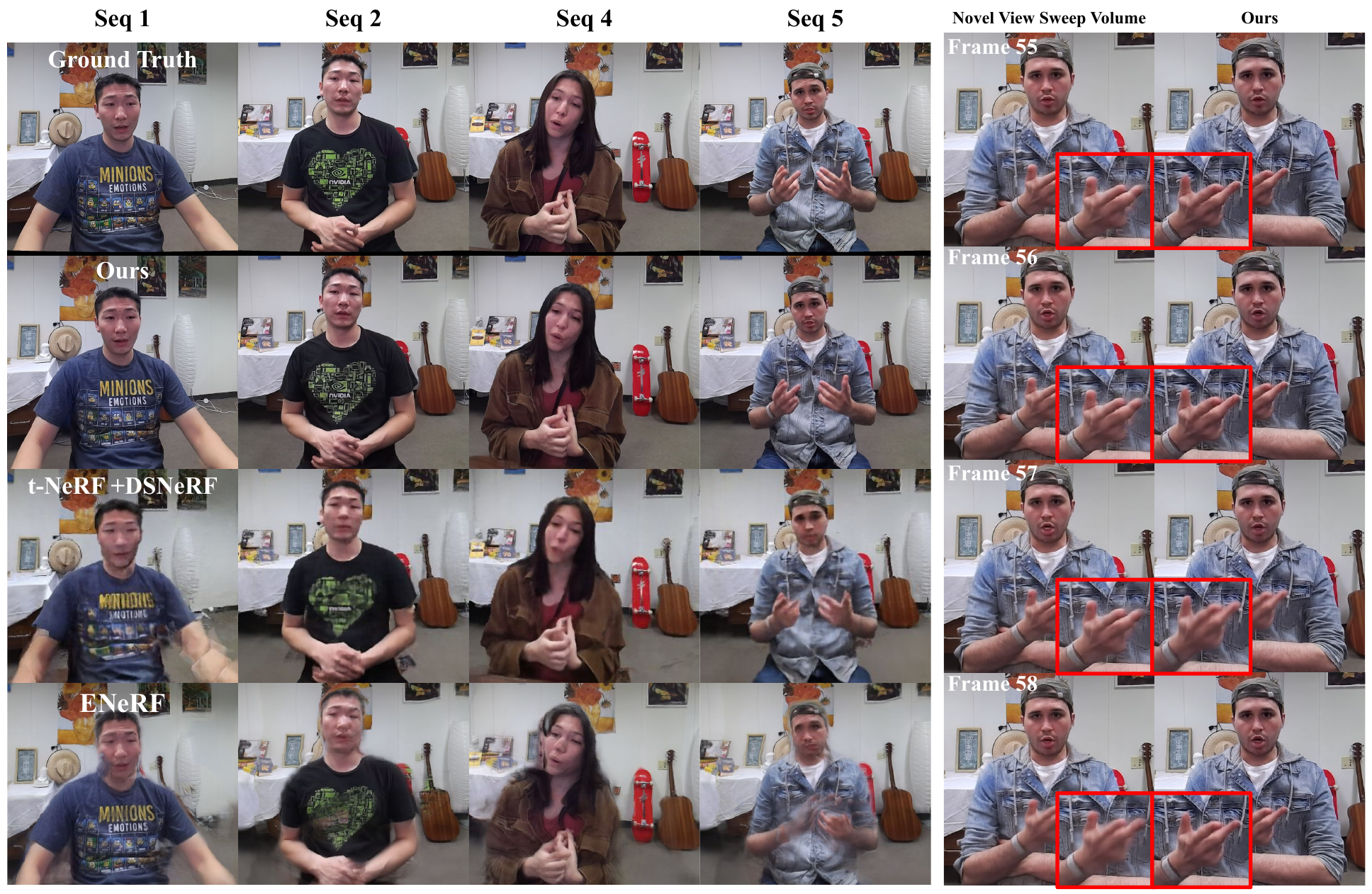}
  \vspace{-1.5em}
   \caption{\small{(a) Comparison with baseline methods. 
   Our neural renderer recovers sharp and fine-grained details in both the foreground and background.  
   (b) Ablation. Compared to conventional depth sweeping from the novel view, our approach shows a more accurate reconstruction of challenging structures like fingers.}}
  \label{fig:viz}
  \vspace{-1em}
\end{figure*}

\section{Training}
The inputs to our system are synchronized RGBD frames from the input cameras, and the output is a free-viewpoint video. Given that there is no existing dataset that uses multiple RGBD cameras to capture video conferencing sessions, we produced our own dataset: the Personal Telepresence Dataset.

\subsection{Personal Telepresence Dataset}
\vspace{-0.5em}
The dataset is captured using the hardware configuration described in Sec.~\ref{sec:hardware}. 
The dataset contains 35 training sequences of a single target user performing various actions in their office. There are also 10 test sequences of various users performing actions in the same office. 
Each sequence contains 4 input RGBD videos and 2 ground truth RGB-only videos, each lasting 20 to 30 seconds. The ground truth cameras are placed near the middle input camera. There are a total of 6 different camera setups (\ie different relative positions and orientations between cameras), creating more diversity in the training data. We also introduce changes to the background decoration to ensure that models can account for variation in the user's environment across sessions. Given that the system is intended for personal use in a specific room, we include only moderate variation to the background. Before each capture session, we capture the background without the user in order to reduce occlusion. Before capturing with a new camera setup, we calibrate the camera poses with a checkerboard and the OpenCV calibration tool. We use the camera intrinsics provided by the manufacturer. 
\subsection{Training Loss}
\vspace{-0.5em}
There are three trainable modules in our rendering network: (1) the image feature encoder $F_\theta(\cdot)$. (2) the density estimation network $E_\omega(\cdot)$, and (3) the temporal renderer $R_\psi(\cdot)$. To save memory, we use a frozen EfficientNetV2s~\cite{EfficientNetv2} pretrained on "ImageNet1K V1"\cite{imagenet} as our encoder $F_\theta(\cdot)$. There is no noticeable impact to performance. We train the rest of the network ($E_\omega(\cdot)$ and $R_\psi(\cdot)$) on our Personal Telepresence Dataset using the following losses:
\begin{equation}
\label{eqn:perc_loss}
L(\Tilde{I}, I) = ||\Tilde{I}-I||_1 + \sum_l \lambda_l ||\phi_l(\Tilde{I})-\phi_l(I)||_1
\end{equation}
\begin{equation}\label{eqn:full_perc}
L_{perc} = L(\Tilde{I}^t, I_{color}^t) +  L(\Tilde{I}^{t*}, I_{color}^{t*})\\
\end{equation}
Eq. \ref{eqn:perc_loss} is the perceptual loss function proposed by\cite{fvsloss}. Eq. \ref{eqn:full_perc} calculates perceptual loss on both low and high-resolution images. $\Tilde{I}^t$ and $I_{color}^{t}$ are the ground truth and rendered image at time $t$. $*$ denotes a high-resolution image.
\begin{equation}\label{eqn:fgd_perc}
L_{fgd\ perc} = L(W\circ \Tilde{I}^t, W\circ I_{color}^t) \\
+  L(W^*\circ \Tilde{I}^{t*}, W^*\circ I_{color}^{t*})\\
\end{equation}
Eq. \ref{eqn:fgd_perc} calculates the perceptual loss on the foreground by masking out the background. $W$ is the foreground mask where a pixel is 1 if its accumulated weight in $W_{fgd}^t$ is greater than $>0.5$, and 0 otherwise. The final loss term is the summation of the general perceptual loss and the foreground perceptual loss: 
\begin{equation}
L_{final}\ = L_{perc} + L_{fgd\ perc}
\end{equation}

\begin{table*}[t]
\centering
\caption{\small{\textbf{Quantitative Comparisons.} \textmd{Values ordered by JOD $\uparrow$ / LPIPS $\downarrow$ / PSNR (db) $\uparrow$ / SSIM $\uparrow$. $\uparrow$ = higher is better, $\downarrow$ = lower is better.}}}
\vspace{-0.5em}
 \resizebox{\textwidth}{!}{
\begin{tabular}{ l c c c c c c c}
    \hline
     & \multicolumn{1}{l}{General} & \multicolumn{1}{l}{Sequence 1} & \multicolumn{1}{l}{Sequence 2} & \multicolumn{1}{l}{Sequence 3} & \multicolumn{1}{l}{Sequence 4} & \multicolumn{1}{l}{Sequence 5} \\
    {\fontsize{11}{104}Methods} & {\fontsize{11}{104}-izable} & {\fontsize{7}{104}\selectfont JOD$\uparrow$ / LPIPS $\downarrow$ / PSNR (db) $\uparrow$ / SSIM $\uparrow$} &  {\fontsize{7}{104}\selectfont JOD$\uparrow$ / LPIPS $\downarrow$ / PSNR (db) $\uparrow$ / SSIM $\uparrow$} &  {\fontsize{7}{104}\selectfont JOD$\uparrow$ / LPIPS $\downarrow$ / PSNR (db) $\uparrow$ / SSIM $\uparrow$} &  {\fontsize{7}{104}\selectfont JOD$\uparrow$ / LPIPS $\downarrow$ / PSNR (db) $\uparrow$ / SSIM $\uparrow$} &  {\fontsize{7}{104}\selectfont JOD$\uparrow$ / LPIPS $\downarrow$ / PSNR (db) $\uparrow$ / SSIM $\uparrow$} \\
    \hline
    \\[-0.8em]
    {\fontsize{11}{104}\selectfont t-NeRF} & {\fontsize{11}{104}\selectfont \xmark} & {\fontsize{11}{104}\selectfont 3.82 / 0.252 / 19.15 / 0.65}  
    & {\fontsize{11}{104}\selectfont 4.70 / 0.112 / 23.11 / 0.80}
    & {\fontsize{11}{104}\selectfont 4.47 / 0.159 / 21.49 / 0.75}
    & {\fontsize{11}{104}\selectfont 3.72 / 0.153 / 18.98 / 0.76}
    & {\fontsize{11}{104}\selectfont 4.67 / 0.128 / 22.60 / 0.79}\\
    \hline
    \\[-0.8em]
    t-NeRF+DSNeRF\cite{dsnerf} & {\fontsize{11}{104}\selectfont\xmark} & {\fontsize{11}{104}\selectfont4.60/ 0.173 / 22.24 / 0.75}& {\fontsize{11}{104}\selectfont4.64 / 0.122 / 23.31 / 0.81}& {\fontsize{11}{104}\selectfont4.50 / 0.146 / 22.35 / 0.77}& {\fontsize{11}{104}\selectfont4.79 / 0.117 / 23.85 / 0.80}& {\fontsize{11}{104}\selectfont4.75 / 0.124 / 23.60 / 0.80}\\
    \hline
    \\[-0.8em]
    {\fontsize{9}{104}\selectfont Gao \etal\cite{dynamicnerf}+DSNeRF} & {\fontsize{11}{104}\selectfont\xmark} & {\fontsize{11}{104}\selectfont3.87 / 0.264 / 20.47 / 0.70}& {\fontsize{11}{104}\selectfont4.27 / 0.194 / 21.22 / 0.75}& {\fontsize{11}{104}\selectfont4.10 / 0.242 / 20.09 / 0.69}& {\fontsize{11}{104}\selectfont4.19 / 0.200 / 21.56 / 0.74}& {\fontsize{11}{104}\selectfont4.02 / 0.283 / 22.24 / 0.74} \\
    \hline
    \\[-0.8em]
    {\fontsize{11}{104}\selectfont
    ENeRF\cite{enerf}} & {\fontsize{11}{104}\selectfont\cmark} & {\fontsize{11}{104}\selectfont4.50 / 0.274 / 17.85 / 0.65} & {\fontsize{11}{104}\selectfont4.24 / 0.284 / 18.01 / 0.66} & {\fontsize{11}{104}\selectfont4.44 / 0.284 / 18.01 / 0.66} & {\fontsize{11}{104}\selectfont4.24 / 0.266 / 18.35 / 0.68} & {\fontsize{11}{104}\selectfont4.56 / 0.266 / 18.35 / 0.68}\\
    \hline
    \\[-0.8em]
    {\fontsize{11}{104}\selectfont Ours Full Model}
    & {\fontsize{11}{104}\selectfont\cmark} & {\fontsize{11}{104}\selectfont\textbf{7.53} / \textbf{0.041} / \textbf{27.92} / \textbf{0.92}} 
    & {\fontsize{11}{104}\selectfont\textbf{7.63} / \textbf{0.037} / \textbf{27.04} / \textbf{0.94}}
    & {\fontsize{11}{104}\selectfont\textbf{7.22} / \textbf{0.061} / \textbf{26.54} / \textbf{0.92}}
    & {\fontsize{11}{104}\selectfont\textbf{7.24} / \textbf{0.050} / \textbf{26.55} / \textbf{0.92}}
    & {\fontsize{11}{104}\selectfont\textbf{7.59} / \textbf{0.042} / \textbf{26.47} / \textbf{0.92}} \\
    \hline
  \end{tabular}
  \label{table:full_results}
 }
  \vspace{-1em}
\end{table*}

\begin{table}[t]
\centering
\caption{\small{\textbf{Ablation Studies} \textmd{Values ordered by JOD $\uparrow$ / LPIPS $\downarrow$ / PSNR (db) $\uparrow$ / SSIM $\uparrow$. $\uparrow$ = higher is better, $\downarrow$ = lower is better. }}}
\vspace{-1em}
 \resizebox{0.49\textwidth}{!}{
\begin{tabular}{ l c c c c}
    \hline
     & \multicolumn{1}{l}{Aaverage on Sequences 1-5} \\
    Methods &  {\fontsize{10}{104}\selectfont JOD $\uparrow$ / LPIPS $\downarrow$ / PSNR (db) $\uparrow$ / SSIM $\uparrow$}\\
    \hline
    Ours Full (MPC+Temporal+SSC) &  \textbf{7.473} / \textbf{0.045} / 26.917 / \textbf{0.925} \\
    \hline
    Ours (MPC+Temporal) &  7.422 / 0.047 / 26.924 / 0.923	 \\
    \hline
    Ours (MPC only) & 7.401 / 0.050 /	\textbf{27.100} / 0.923 \\
    \hline
    Novel View Sweep Volume & 7.349 / 0.053 / 26.946 / 0.919\\
    \hline
  \end{tabular}
  \label{table:ablation}
 }
 \vspace{-0.5em}
\end{table}

\begin{table}[t]
\centering
\caption{\small{\textbf{Description of Test Sequences} \textmd{The term "\textbf{same}" refers to something that is part of the training data. The term "\textbf{new}" refers to something that did not appear in the training data and is thus new to the model. All sequences except for "Sequence 1" is evaluated from new novel viewpoints. This enables a rough understanding of how the performance changes when the novel viewpoint is not covered by training data.}}}
\vspace{-0.5em}
 \resizebox{0.45\textwidth}{!}{
\begin{tabular}{ l c c c c c}
    \hline
     & \multicolumn{1}{l}{User} & \multicolumn{1}{l}{Clothing} & \multicolumn{1}{l}{Input Views} & \multicolumn{1}{l}{Evaluation Views} \\
     \hline
     Sequence 1 & Same & New & Same & Same\\
     \hline
     Sequence 2 & Same & New & New  & New\\
     \hline
     Sequence 3 & New & New & New  & New\\
     \hline
     Sequence 4 & New & New & New  & New\\
     \hline
     Sequence 5 & New & New & New  & New\\
     \hline
  \end{tabular}
  \vspace{-1.5em}
  \label{table:test_setting}
  
 }\vspace{-3mm}
\end{table}

% \vspace{-5mm}
\subsection{Variants}
\vspace{-0.5em}
\textbf{Our Full Model} uses MPC, the temporal renderer and Spatial Skip Connection (SSC). It is trained in two stages in order to save memory: 
Stage (1): the model (\ie $E_\omega(\cdot)$ and $R_\psi(\cdot)$) is first trained for 12 epochs to generate a novel view image for a single frame. Stage (2): With the density network $E_\omega(\cdot)$ trained and frozen, the renderer $R_\psi(\cdot)$ is re-initialized and re-trained for 6 epochs to produce 2 novel view images at the same viewpoint from 2 frames of input. In this way, the training memory consumption at Stage 2 is greatly reduced.

We also evaluate multiple variants: \textbf{Ours(MPC only)}, \textbf{Ours(MPC+Temporal)}, and \textbf{Novel View Sweep Volume}(no temp, SSC, MPC). These variants omit various components for ablation purposes, and they are trained with Stage 1 only. 
Note that the \textbf{Novel View Sweep} variant uses conventional cost volumes, \ie depth sweep from the novel view camera. Similarly to VirtualCube, we first lift input depth maps from all views into point clouds, which are then rasterized into novel view depth maps. The average of these depth maps are then used as the center of the local depth sweep. For a fair comparison, we use the same number (\ie 12) of sweeps with a step size of 1cm (\ie between -6cm and +6cm of the average depth map). 

Notice that all variants use the \textbf{same pre-captured background} as our full model and thus the \textbf{same background feature image} during alpha blending, resulting in very similar performances on the background.

\noindent \textbf{Implementation Details} We train the network on 4 NVidia RTX 3090 GPUs using the AdamW~\cite{adamw} optimizer with a learning rate of $10^{-4}$, and a batch size of 4. To reduce memory consumption during training, we render random crops of size $384\times512$ for the low-resolution synthesis and $768\times1024$ for high-resolution. During testing and evaluation, the low-resolution rendering is of size $480\times640$ and the high-resolution rendering is of size $960\times1280$. We preprocess MPC depth volumes and store them locally. While there are many real-time depth-map/point-cloud renderers, we use Pytorch3D\cite{pytorch3d} to render the depth maps due to its simplicity.

\section{Evaluation}
Our model is trained on a single male target user in his office using 4 RGBD cameras as inputs and 2 RGB-only cameras as ground truth supervision. During testing and evaluation, performances on the 2 ground truth cameras are averaged to provide a more robust assessment. 

\textbf{Evaluation Metrics} We use LPIPS~\cite{LPIPS}, SSIM, and PSNR to measure per-frame image quality. We also use JOD~\cite{JOD} as an important metric to evaluate video stability.

\textbf{Test Sequences} We also consider the different factors that may affect the evaluation results: (1) whether or not the novel viewpoints are included in the training data, (2) robustness to new clothing, (3) robustness to new users. Therefore, we employ different settings for the 5 test sequences as shown in Table.\ref{table:test_setting}. We also include results on 5 additional test sequences in our supplementary materials. Since this is a personal system, we do not introduce significant differences to the environment during testing.

\subsection{Baselines}
\vspace{-0.5em}

Our goal is to generate free-viewpoint videos from a few RGBD cameras in a video conferencing setup. Given that few competitive baselines are specifically designed for this application, we adapt recent neural rendering methods to this application in order to study a wider spectrum of possible solutions to the problem. These baselines include: 
(1) \textbf{t-NeRF}: NeRF~\cite{nerf} with time as an additional input, (2) \textbf{Gao \etal~\cite{dynamicnerf}}: models the scene with neural radiance and scene-flow fields. We also enhance baselines (1) and (2) with DSNeRF~\cite{dsnerf}'s depth loss term, which is a KL divergence term used to encourage uni-modal density distribution near the input depth value.
(3) \textbf{ENeRF~\cite{enerf}}: a real-time generalizable radiance field optimized with a plane-sweep.
(4) \textbf{Novel View Sweep Volume} (Tab.~\ref{table:ablation}): Similar to Microsoft VirtualCube~\cite{virtualcube} (code not available), this baseline generates cost volumes via depth sweeps from the novel view camera. This is a variant of our system without MPC, SSC, or temporal rendering.

\subsection{Results and Comparison}
\vspace{-0.5em}
\textbf{Quantitative Results.} As shown in Table. \ref{table:full_results}, our system outperforms all NeRF-based approaches in test sequences 1-5 by a large margin; this is likely due to the sparsity of viewpoints. Moreover, our system maintains a high quality even when applied to "new" users that are not in the training data (Sequences 3-4). Additionally, Sequences 1 and 2 achieve similar results despite different camera setups (\ie input and ground truth camera poses), implying that our system performs well with new viewpoints.

\textbf{Qualitative Results.} Fig. \ref{fig:viz} shows novel view renderings for various sequences. Our system is robust to sparse viewpoints, reconstructs complex hand gestures with high fidelity, and generalizes well to new users, whereas our baselines struggle with sparse viewpoints. While it is not possible to directly compare with VirtualCube~\cite{virtualcube}, we show in supplementary videos that our system renders more stable videos and is more robust to large fast movements.

\textbf{Ablation Studies.} 
Table.~\ref{table:ablation} shows the performance of different variants of averaged over the 5 test sequences. Note that the \textit{pre-captured background is shared across all models}. This implies very similar results on a large portion of the images, and that the quantitative improvements mostly arise from the foreground.
Despite sharing the backgrounds, our proposed modules still show consistent improvements in accuracy (LPIPS, SSIM) and stability (JOD). 
In the supplementary videos, we provide clear visualizations of improvements in the foreground area. The most significant improvement comes from our MPC volume, and our temporal renderer and Spatial Skip Connection further improve the stability and accuracy.

% \vspace{-0.5em}
\section{Limitation and Future Work}  
\vspace{-0.5em}
(1) Although our rendering system is fast, it is not real-time yet. Currently, the construction of cost volumes consumes the majority of the rendering time. To achieve real-time rendering, we will experiment with smaller volumes, sparse volumes, sparse convolution, and lighter networks. (2) Our current system does not render free-viewpoint videos via immersive display technologies (\eg autostereo displays). In the future, we will track the user's pupil positions and leverage autostereo displays to render separate videos for each eye to create depth cues essential for immersive telepresence. 

% \vspace{-0.5em}
\section{Conclusion}  
\vspace{-0.5em}
We presented a capture and rendering system designed for personal telepresence systems. Our capturing system only requires a few RGBD cameras and thus can be easily installed on typical desks, unlike recent commercial systems. Our rendering system renders high-quality free-viewpoint videos, and it outperforms recent view synthesis methods in terms of both accuracy and stability. It can accurately reconstruct complex hand gestures, fast body movements, and rich environmental details without object templates, complex pre-processing, or costly optimization.  

By proposing a personal system, we work to democratize immersive telepresence experiences. The lightweight nature of this setup facilitates more intimate and informal connections between remote individuals without requiring the resources needed by commercial counterparts. As such, our system helps increase access to telepresence for the general public and encourages future work in this area.

%%%%%%%%% BODY TEXT

{\small
\bibliographystyle{ieee_fullname}
\bibliography{egbib}
}

\end{document}